\documentclass[10pt,twocolumn,letterpaper]{article}

\usepackage[pagenumbers]{cvpr}

\definecolor{cvprblue}{rgb}{0.21,0.49,0.74}
\usepackage[pagebackref,breaklinks,colorlinks,allcolors=cvprblue]{hyperref}

\def\paperID{17482} 
\def\confName{CVPR}
\def\confYear{2026}

\title{Advancing Digital Twin Generation Through a Novel Simulation Framework and Quantitative Benchmarking}

\author{Jacob Rubinstein, Avi Donaty, Don Engel\\
University of Maryland, Baltimore County (UMBC)\\
Baltimore, Maryland, USA\\
{\tt\small \{jrubins1, adonaty1, donengel\} @umbc.edu}}

\begin{document}
\maketitle

\begin{abstract}
The generation of 3D models from real-world objects has often been accomplished through photogrammetry, i.e., by taking 2D photos from a variety of perspectives and then triangulating matched point-based features to create a textured mesh. Many design choices exist within this framework for the generation of digital twins, and differences between such approaches are largely judged qualitatively. Here, we present and test a novel pipeline for generating synthetic images from high-quality 3D models and programmatically generated camera poses. This enables a wide variety of repeatable, quantifiable experiments which can compare ground-truth knowledge of virtual camera parameters and of virtual objects against the reconstructed estimations of those perspectives and subjects. 

\end{abstract}

\section{Introduction}
\label{sec:intro}

Some of the oldest methods for the creation of digital twins of 3D real-world objects involve the hand-crafting of models, with engineers using CAD software and visual artists using tools like Blender. Some of the newest methods involve generative AI, including models driven by linguistic prompts and/or single-shot 2D images. Despite the promise of these traditional and emerging models, multi-view photogrammetry remains a very popular tool for digitizing the 3D world. 

Each stage of the photogrammetry pipeline involves considerable design choices. In choosing or setting the scene to be captured, one might have control over lighting (by time of day or artificial lighting) and backdrop. In taking photos, one might have the choice of a camera and lens, and one must certainly choose how many photos to capture and the position and orientation of the camera(s) for each photo. In processing the photos, a particular feature detection algorithm (SIFT \cite{sift}, SURF \cite{surf}, etc) must be selected in order to generate point-based features for matching. In matching those points, thresholds must be set for what constitutes a matched pair and an appropriate number of total pairings. In triangulating these points, constructing a 3D model (e.g., mesh), and assigning color and other surface properties (e.g., a texture for a mesh), additional parameters must be specified. Analogous choices exist across all sorts of 3D representations (mesh, splat \cite{splat}, NeRF \cite{nerf}, etc.). Prepackaged software solutions may offer defaults, but experienced users will develop opinions, based on qualitative experiences, on which parameters to override, under which circumstances, to achieve desirable digital twins.

By simulating a scene and creating synthetic photos from a simulated scene and virtual cameras - where ground truth is readily available - it becomes possible to replace qualitative judgment with repeatable, quantifiable experiments. The same synthetic data can also be used not only for improvements to photogrammetry itself, but also to create repeatable experiments in pose estimation (e.g., for SLAM \cite{slam}) and to train generative AI models (for photogrammetry, SLAM, etc.). In this paper, we focus on the profound implications of this approach for digital twin generation, demonstrating how our novel pipeline enables the precise comparison of reconstructed objects and camera poses against their known virtual ground truths, offering a powerful new methodology for improving the generation of high-fidelity 3D models.

\section{Related Work}
\label{sec:relatedWork}

\subsection{Synthetic Data for Computer Vision and Computer Graphics}

The use of synthetic data has become a pivotal technique in the development and testing of modern intelligent systems. Artificially generated datasets are used in training deep neural networks for tasks in computer vision and image generation \cite{jimaging8110310, paulin2023review}. In one recent example \cite{360photos}, synthetic indoor scenes were captured with a virtual 360{\textdegree} RGB-D camera to train a model which could then take flat 360\textdegree content and convert it to a Gaussian splat representation for use in virtual reality. 

The use of synthetic data has proven particularly useful in developing autonomous robotics \cite{dimitropoulos2022brief} and self-driving vehicles \cite{hu2023simulation}, where ``sim2real'' strategies leverage simulated physics and virtual environments to train robust AI models prepared to navigate the physical world. The power of simulation also extends to complex social dynamics; for example, synthetic photos have been used to test the sensitivity to camera perspective of vision-language models \cite{aixvr}; and virtual reality has been used to capture simulated human-robot interactions \cite{simulated-HRI}, enabling the creation of training data for robots designed to operate safely and effectively alongside people.

\subsection{Virtual Photogrammetry}
In a proof of concept project showing the promise of synthetic data for photogrammetry \cite{virtual-photogrammetry}, Esmaeili and Thwaites captured 2D screenshots of a video game (World of Tanks) and of screen-based renderings of photogrammetrically captured 3D models. They entered these images into commercial photogrammetry software (AgiSoft PhotoScan) and provided a qualitative analysis of the resultant reconstructions of the original 3D models. This succeeded in its aim to show the feasibility and limitations of such reconstruction. Relative to our work, this approach was not well-positioned to create repeatable or quantitative experiments because their methodology lacked programmatic control over their camera parameters and of the contents of their scenes. Another project, ``DigiDogs,'' \cite{digiDogs} also used screen captures from a video game as synthetic data. In this case, they used images of dogs from Grand Theft Auto V in order to build a pose estimation model for dogs in the real world.

While those projects depended on renderings from third-party games or software, other prior work has coupled the generation of novel 3D scenes with the simulation of the photogrammetric reconstruction of those same scenes, providing those authors with the control and ground-truth knowledge required to ask quantitative questions. Prior work of this sort has focused on improving specific technologies aimed at particular use cases. 

A series of research projects by Piatti et al \cite{piatti2006virtual, piatti2013virtual} have focused on simulating aerial photogrammetry, with motivation coming from the desire to optimize mission planning, improve data acquisition strategies, and ultimately reduce the costs associated with airborne surveys. In alignment with these goals, these projects used simplified 3D scenes and basic pinhole cameras, which limit the relevance to using synthetic data to improve the quality of digital twin creation through photogrammetric design choices.

Similarly, simulated photogrammetry has been used to model the detection of deformities in the Sardinia Radio Telescope \cite{buffa2016simulation}; to explore novel photogrammetric approaches for target objects that are cylindrically shaped \cite{kortaberria2019theoretical}; and to improve the design of photogrammetry rigs meant specifically for ``human body scanning and avatar creation'' \cite{gajic2019simulation}. Our work differs from these in that we aim to provide a generalizable framework for advancing many types of digital twin generation and pose estimation, with any type of object or scene.

We have identified one project, by Mezhenin et al \cite{mezhenin2020using}, which shared our aim of building a synthetic data pipeline from ground truth digital twins in order to improve digital twin generation in general. Their project was focused on benchmarking the relative performance of different commercial photogrammetry products and focused specifically on comparing point clouds (using the Hausdorff metric), whereas we introduce tools and methodology for running a broader set of experiments on the many parameters involved in a photogrammetric pipeline.

\section{Methodology}
\label{sec:methodology}
This section proposes a general methodology for preparing photogrammetric experiments. We then, in Section \ref{sec:currentApproach}, detail our current implementation of this methodology.

\subsection{Ground Truth Pose Generation}
Photogrammetry sometimes uses the same physical camera repeatedly to generate multiple images and sometimes uses many cameras (e.g., in a rig) to capture many images simultaneously. Here, we refer to ``cameras'' such that they have a one-to-one relationship with each image. That is, a camera taking two photos for the same reconstruction would be considered as though it were two cameras, each with their own pose and intrinsics (e.g., the zoom may have changed from one image to the next).

There are a wide variety of approaches to choosing camera poses (position and orientation) for 3D reconstruction.  The number of options increases when capturing a full scene relative to capturing a single object. For the purpose of this work, we narrow our domain to captures of small single objects. 

Even in this narrowed domain, there are many factors to consider. The distance of the camera to the object and the spacing of cameras will have direct impact on the quality of the resultant digital twin. It seems likely that an object with small protrusions or crevasses will require cameras to be well placed such that they can capture the parallax of these details shifting from one perspective to another. However, there is likely to be a point of diminishing returns beyond which additional coverage will add more computational complexity without adding a commensurate improvement in the quality of the resultant model.

In preparing experiments to place and pose cameras, a variety of spatial patterns could be used, or the placement (and quantity) could be algorithmically derived from the subject itself. 

\subsection{Render Frames}
In addition to deciding on camera poses and extrinsic parameters (locations, orientations), the camera intrinsics (i.e., focal length and other internals) must be determined. Similarly, aspects of the scene may involve choices, such as lighting sources and the background (i.e., choosing a color or an image). After rendering the simulated frames, an optional additional step involves appending the camera intrinsic parameters, such as camera make, model, focal length, and sensor width, as metadata to the rendered images files. Doing so allows for reconstruction software to avoid estimating these factors, which is likely to affect the quality of the final reconstruction. 

\subsection{3D Reconstruction}
We then take these simulated frames as input for 3D reconstruction. There are several forms this reconstruction could take, such as triangular meshes, NeRFs, or 3D Gaussian Splats. The requirement of this step is taking the rendered frames as input and outputting a 3D representation of the object, which can be compared to the starting representation. Choices made at this point in the pipeline may include which algorithms to use (e.g., choice of feature detection techniques), which parameters or thresholds to use in the reconstruction process (e.g., confidence thresholds for feature matching, density settings for point cloud generation, or filtering parameters for outlier removal), and the choice of camera models to solve for during Structure from Motion (SfM). Each of these choices can influence the fidelity and accuracy of the final 3D model, and by analyzing them in a controlled virtual environment, we can better understand their impact on the resulting reconstruction.

\subsection{Analyze Results}
There are many possible approaches to comparing the starting object with the reconstructed one, each suiting different goals. The choices made in earlier stages of the pipeline, such as the choice of 3D representation, also play a large role in this step. Objects can be compared to each other in a variety of ways. Volumes could be compared directly by aligning them and to find the minimized Intersection over Union (i.e., Jaccard index). 2D images, captured from pairs of matching perspectives, could be compared using Peak Signal-to-Noise Ratio (PSNR) or Structural Similarity Index (SSIM). Point clouds (i.e., the matched features) could be compared directly using Hausdorff distance \cite{mezhenin2020using} or Chamfer distance. In addition to the quality of the results, the computational costs (including runtime) may contribute to an analysis.

\section{Current Approach}
This section details a specific initial implementation of our methodology described in Section \ref{sec:methodology}, including the tools we used as well as the design decisions made and the explanations of those decisions. To demonstrate this approach, we used the dataset published with ``A Real World Dataset for Multi-view 3D Reconstruction'' \cite{real_world_dataset}. This set of detailed digital twins, which its authors captured using high-end 3D scanners, exists in order to support reproducible photogrammetry research.

\label{sec:currentApproach}
\subsection{Ground Truth Pose Generation}
\label{subsec:poseGeneration}
We took a simple approach to determine the set of camera poses, arranging the cameras evenly in a sphere centered at the object's origin. This method takes advantage of the synthetic environment allowing a clear view of all angles of the model, maximizing overlap between neighboring frames. We decided to take this approach to ensuring feature matching will have targets and providing views of every outward face of the object.

\begin{figure}[htbp]
\centerline{\includegraphics[width=\linewidth]{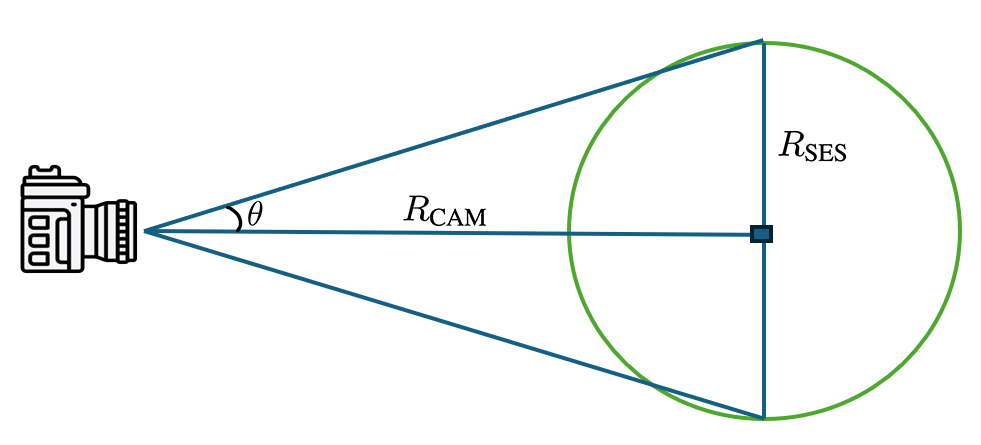}}
\caption{In green: The SES found using Welzl's algorithm. $R_{\textrm{SES}}$ is the radius of the smallest enclosing sphere. $\theta$ is half of Blender's default vertical field of view. $R_{\textrm{CAM}}$ is the radius of the camera sphere (what we're trying to find).}
\label{fig:SES-Diagram}
\end{figure}

To determine the center of the object we use Welzl's algorithm \cite{welzl}, a recursive algorithm that finds the smallest enclosing sphere (SES) around a set of points. We utilize an implementation of Welzl's algorithm \cite{WelzlsAlgorithmGitHub} that calculates the SES and paramatrizes it by the center and radius of the sphere. We then calculate the radius of the camera sphere by using the center and radius of the SES and Blender's default vertical field of view (roughly \( 23^\circ \)) to form the basis of a right triangle (Figure \ref{fig:SES-Diagram}). We derive the camera radius as follows:

$$\tan(\theta) = \frac{R_{\textrm{SES}}}{R_{\textrm{CAM}}}$$ 
$$R_{\textrm{CAM}} \tan(\theta) = R_{\textrm{SES}}$$ 
$$R_{\textrm{CAM}} = \frac{R_{\textrm{SES}}}{\tan(\theta)}$$ \\

\normalsize With the camera sphere's radius determined, we generate 100 roughly evenly spaced points using a Fibonacci sphere distribution \cite{fibSphere}, randomizing the order of the poses to avoid any bias introduced by this method of generation. We use 100 camera poses, as this provides a good middle ground between accuracy and computation time.

\subsection{Render Frames}
\label{subsec:renderFrames}
With a list of set camera poses, the next step is rendering synthetic photographs of the 3D model; we use Blender \cite{blender} and its Python API for this task. We first clear the scene of all starting object and then import our 3D model, including all associated textures. Next, the object's orgin is set to its center of mass, and then the location is set to (0, 0, 0).

With the model added, we need to add both lights and a camera to render the scene. A camera is added to the scene with a \textit{DAMPED\_TRACK} constraint who's target is our model and who's tracking axis is set to \textit{TRACK\_NEGATIVE\_Z}. This constraint forces the pitch and yaw of the camera to direct at the our model, while allowing control over the roll. To ensure even lighting of the model, we create a cube of 6 area lights, each directed at the origin. These lights form an enclosed cube of even lighting from all directions.

The scene is now ready for rendering, so we take the generated camera poses and store the location and rotation values as successive keyframes by the camera. The roll of the camera is also set to a random angle between 0$^{\circ}$ and 360$^{\circ}$, this is done to reduce any orientation bias which might be introduced into reconstruction. The series of camera poses are then exported as an alembic file, allowing for future comparison to the reconstructed camera poses.

The final steps of the rendering process are setting the background of the scene to a solid color, the resolution of the scene to 2560 x 1440, the render percentage to 100\%, and turning off denoising. We then step through the camera poses and render a JPEG image of the scene for each.

\subsection{3D Reconstruction}
The next step in the pipeline is the process of taking the 2D frames rendered in the previous step and using them to create a 3D model. To accomplish this, we use the photogrammetry pipeline AliceVision Meshroom \cite{meshroom}. 

Since we want to perform experiments in batches, we need a command-line interface. We utilize Meshroom CLI \cite{meshroom_cli} for automation of the process.

Meshroom uses a nodal system in which each node is a specific operation, and the output of one node can be the input of a future node. The attributes of each node can be customized to better serve the needs of a specific project.

Meshroom CLI uses a Python script to call each of the nodes used by Meshroom. The main function of the script calls all the nodes, creating folders to hold the outputs of each of the steps.

The first step Meshroom uses is \textit{Feature Extraction}. This is identifying groups of pixels in the frames that are distinctive from multiple viewpoints. Each group found is considered a feature. 

Next is \textit{Feature Matching}. This is the process of finding corresponding features across multiple images of the same scene. These matched features are vital for the next step.

\textit{Structure From Motion} uses the matched features to compute the relative position and orientation of the cameras that captured those images. The 3D coordinates of the matched features are also triangulated to create a point cloud, a sparse representation of the scene. We now have accurate camera poses. This step is the foundation for the later steps where a dense reconstruction is created.

Next is \textit{Depth Map Estimation}, where we generate a depth map for each image, estimating the distance from the camera to the object surface (i.e give each pixel in the image a depth value). We then use these maps for dense reconstruction, also known as \textit{Meshing}. A dense point cloud is created from the depth maps and that dense point cloud is converted into a triangular mesh. We then apply the textures from the original frames onto the mesh to create a realistic 3D model. 

The final folder contains the Wavefront OBJ file, the JPEG texture file, and an MTL file that specifies the material properties of the model, such as the ambient, diffuse, and specular colors.

One thing we added to the Meshroom CLI script was to call an additional node, \textit{Export Animated Camera}. This node takes as input the results from the Structure From Motion step, and exports the estimated camera positions and rotations as an alembic file similar to the file created in the image generation step. This additional step enables us to compute a rough rigid transformation between the ground truth and estimated camera poses. We can then transform the reconstructed model the same way, which is a first step towards exactly aligning the models.


\subsection{Analyze Results}
We use a weighted structural similarity (SSIM) \cite{SSIM} measure as a quantitative value to analyze the quality of the reconstructed models. For each reconstructed model, we apply a transformation which aligns the mesh closely in scale, rotation, and translation to the mesh of the ground truth model. With the models aligned, we can render frames of the reconstructed model using the same poses and rendering settings as were used for the ground truth rendered frames. We then compute our weighted SSIM between the corresponding pairs of ground truth and reconstructed renders, and then average these 100 individual SSIM values to achieve our final score for the quality of the reconstruction.

\subsubsection{Alignment}

We use the iterative closest point (ICP) \cite{ICP} algorithm to calculate the transformation between the reconstructed and ground truth models. This algorithm iteratively moves one model, in this case the reconstructed model, to minimize the distance between the distance between the points in the moving model and the static model, in this case the ground truth model. ICP can struggle when the two models differ significantly in position, rotation, or scale, so we utilize an additional rough alignment transformation.

To achieve this rough alignment, we use the position and rotation information stored in the alembic files to calculate a transformation which best aligns the set of poses estimated by Meshroom to the known ground truth set of poses. This transformation can then be applied to the reconstructed model, resulting in a pose which is roughly aligned to the ground truth model.

The two sets of camera poses are corresponding pairs, as Meshroom exports the camera poses in the same order as the renders were taken. We can leverage this fact for aligning the poses by treating the camera positions as 3D vectors and using the Kabasch algorithm, as implemented in the SciPy \cite{scipy} $align\_vectors$ function, to calculate the optimal rotation matrix that best aligns the paired sets of points. Before calculating the rotation matrix, we must first align the translation and scale of the set of reconstructed poses.

 Let \( P_{\text{est}} = \{ \mathbf{p}_{est\_1}, \mathbf{p}_{est\_2}, \dots, \mathbf{p}_{est\_n} \} \) be the set of estimated camera poses, and let \( P_{\text{gt}} = \{ \mathbf{p}_{gt\_1}, \mathbf{p}_{gt\_2}, \dots, \mathbf{p}_{gt\_n} \} \) be the set of ground truth camera poses. Define the center of each set as:

\[
\mathbf{c}_{\text{est}} = \frac{1}{n} \sum_{i=1}^{n} \mathbf{p}_{est\_i}, \quad \mathbf{c}_{\text{gt}} = \frac{1}{n} \sum_{i=1}^{n} \mathbf{p}_{gt\_i}.
\]

To remove the translation introduced by Meshroom, we compute a translation factor as:

\[
t = \mathbf{c}_{\text{est}} - \mathbf{c}_{\text{gt}}
\]

We compute the translated set of camera poses \( P'_{\text{est}} \) by subtracting the translation factor from each pose:
\[
P'_{\text{est}} = \left\{ \mathbf{p}'_{est\_i} = \mathbf{p}_{est\_i} - t \mid i = 1, 2, \dots, n \right\}.
\]

Next, we compute the scaling factor \( s \) to eliminate scaling differences. The scaling factor is given by:

\[
s = \frac{\frac{1}{n} \sum_{i=1}^{n} \|\mathbf{p}_{gt\_i} - \mathbf{c}_{\text{gt}}\|}{\frac{1}{n} \sum_{i=1}^{n} \|\mathbf{p}'_{est\_i} - \mathbf{c}_{\text{gt}}\|}.
\]

We then take \( s \) and apply a matrix scaling transformation to \( P'_{\text{est}} \), resulting in \( P''_{\text{est}} \). Finally, the scaled and translated set \( P''_{\text{est}} \) is used as input to the  $align\_vectors$ function along with the ground truth set \( P_{\text{gt}} \). We apply these three transformations to the reconstructed model to achieve our rough alignment.

With our reconstructed model now roughly aligned to the ground truth one, we are free to utilize ICP to minimize the point-wise distance between the models. For this, we utilized the 
ICP (Iterative Closest Point) Registration / Alignment Blender add-on \cite{ICP_Tool}.

\subsubsection{Comparison Metric}

To quantify the accuracy of the reconstruction, we use the SSIM measure from the scikit-image library \cite{van2014scikit}. However, there are some issues when using it in its default form. 

Our images are represented as $x \times y \times 3$ arrays, corresponding to height, width, and color channels. The default SSIM uses a convolutional approach over each pixel in the corresponding  images.

For each 11x11 window, a weighted combination of similarities for luminance, contrast, and structure is computed as the SSIM for pixel ($i, j$) - the center pixel of the window. The SSIM score has a range of (-1, 1), where 1 indicates perfect similarity, 0 indicates no similarity, and -1 indicates a perfect anti-correlation. This similarity map has a shape of $x \times y$. At the end, it takes the average of all the per-pixel SSIM scores and outputs a single value that is the total similarity score between the images.

Using this set up, the background is weighted just as heavily as the foreground. However, the background of each of our images is the same solid color and occupies a large percentage of the frame. Therefore, the default SSIM score for our images will generally lean towards positive even if the reconstruction is poor.

We resolve this by setting the weight for the per-pixel SSIM values of the background to 0. Since we set the background color to a specific RGB value, we can get the indices of the pixels where both images have the background RGB value. We then set the weight of those indices of the similarity map to 0, and everything else to 1. When taking the weighted average, we now get the average of per-pixel similarities that only correspond to the model itself.

We do this for each corresponding pair of frames of the models, and the average weighted SSIM score between the frames is our global score for the model.

\section{Results}
\label{sec:results}

\begin{figure}[htbp]
    \centering
    {\label{fig.a}\includegraphics[width=\linewidth]{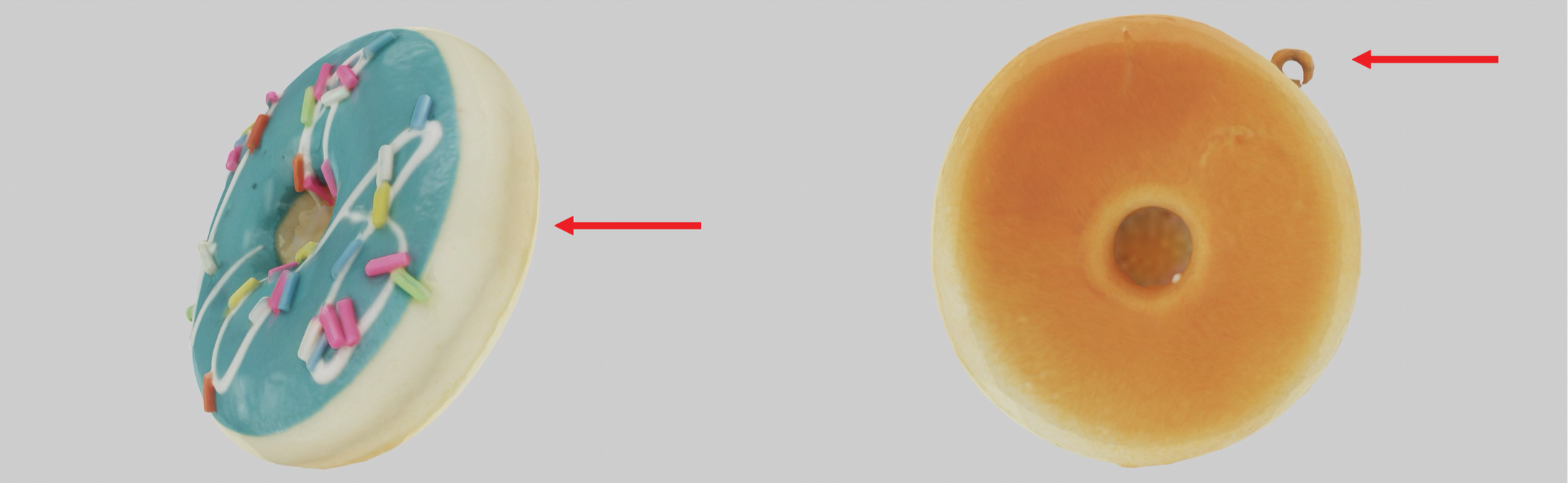}} \\
    {\label{fig.a}\includegraphics[width=\linewidth]{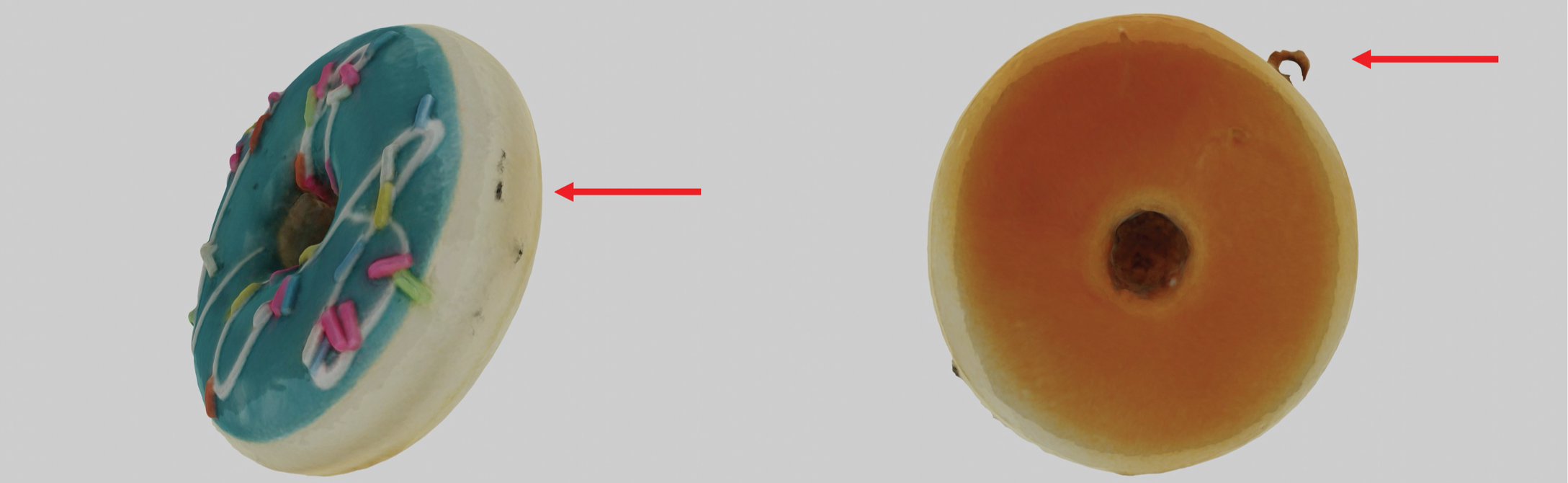}} \\
    
    \vspace{0.1in}
     
    \subfloat{\label{fig.b}\includegraphics[width=\linewidth]{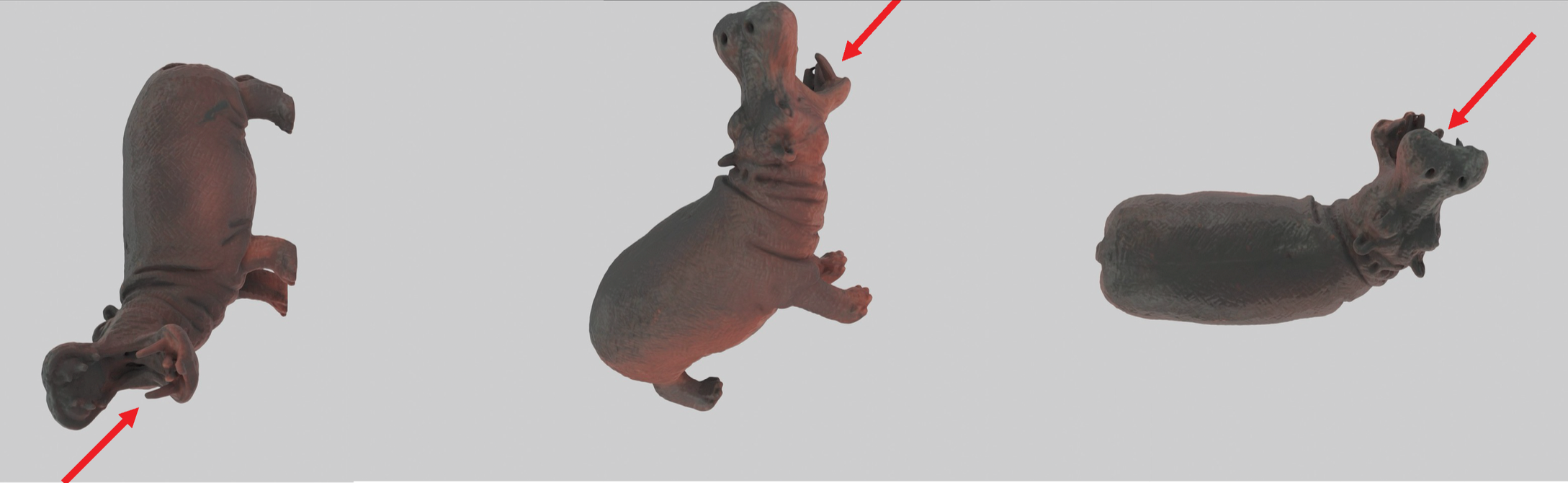}} \\
    \subfloat{\label{fig.b}\includegraphics[width=\linewidth]{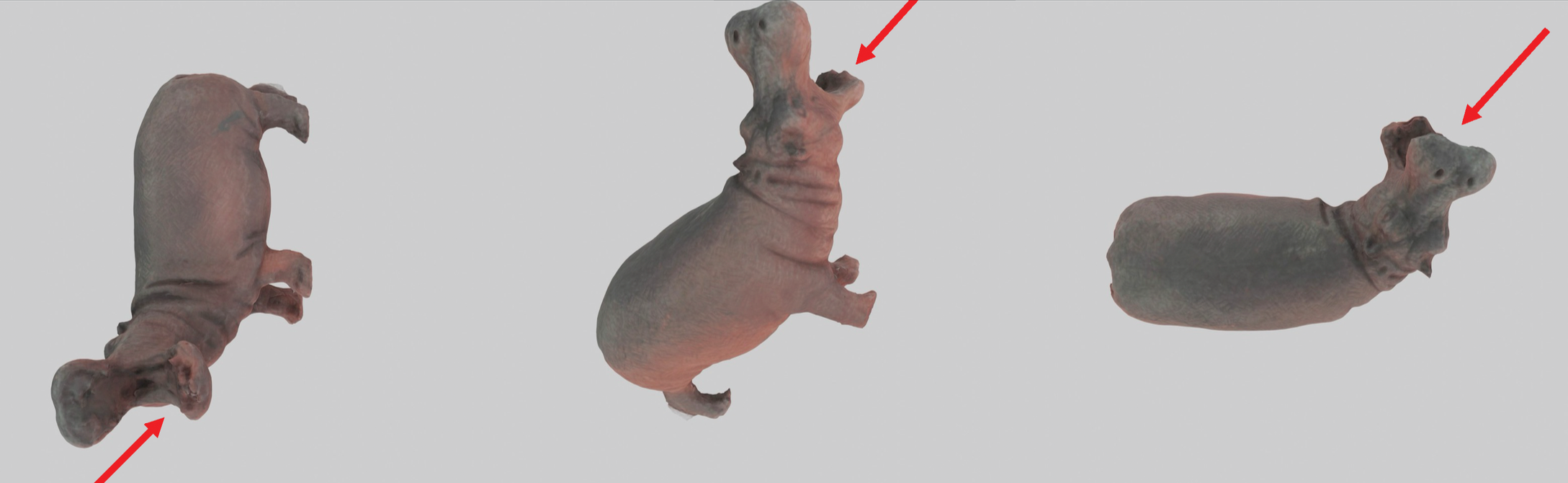}}
    \caption{Original models, followed by reconstructed models, using the same camera positions. Overall, the reconstructions were similar to the original models, but certain common model traits caused issues. Red arrows demonstrate differences.}\label{fig.global}
    \label{donut-and-hippo}
\end{figure}

Of the 805 models, 688 (85.5\%) successfully completed all stages of the pipeline, including the final ICP alignment step. The remaining 117 models failed to produce a final output.

\begin{figure}[htbp]
    \centering
    \includegraphics[width=\linewidth]{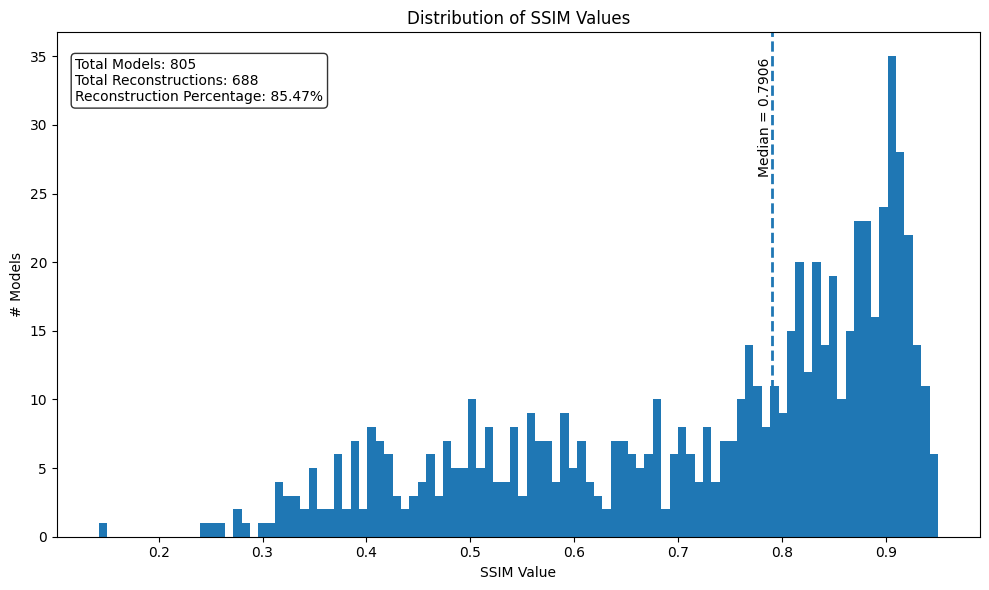}
    \caption{Distribution of SSIM values across the entire dataset. The x-axis represents the SSIM score, the y-axis represents the number of models with that score.}
    \label{fig:histogram}
\end{figure}

For the successfully reconstructed models, we observed a wide spectrum of quality. As illustrated in Figure \ref{donut-and-hippo}, many reconstructions captured the overall shape and color of the original objects but suffered from a loss of fine detail. For instance, the loop on the side of the toy donut model was not resolved, and the texture appeared smoothed and darkened. Similarly, the hippo model, while recognizable, lost definition in its facial features and the texture was brighter than in the original. In other examples, a common artifact was the inclusion of the background color into the reconstructed texture, particularly on object edges, which diluted the texture's color fidelity.

\begin{figure}[htbp]
    \centering
    {\label{fig.a}\includegraphics[width=\linewidth]{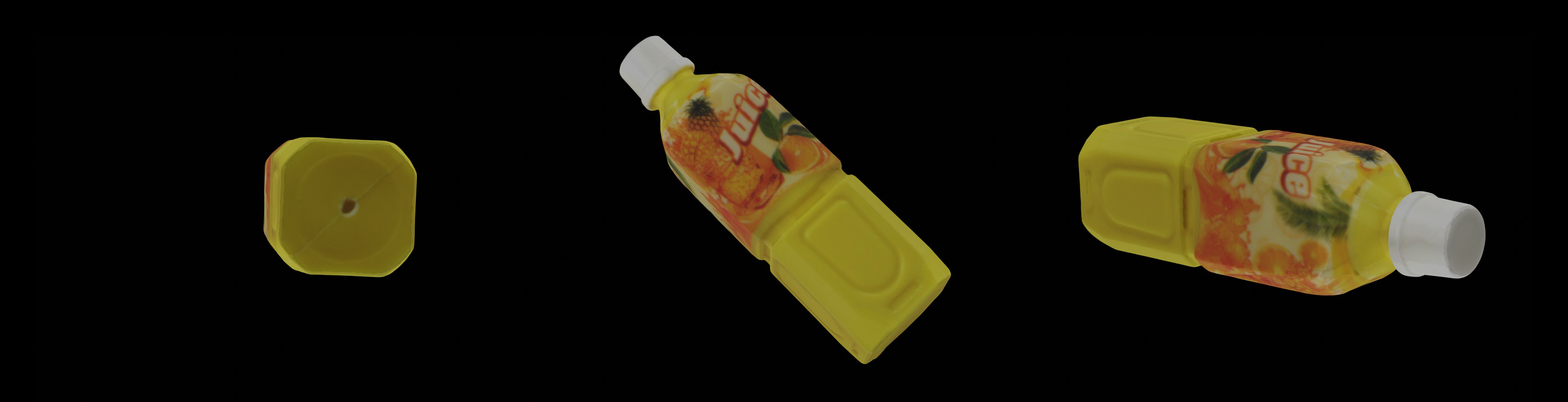}} \\

        \vspace{0.1in}

    {\label{fig.a}\includegraphics[width=\linewidth]{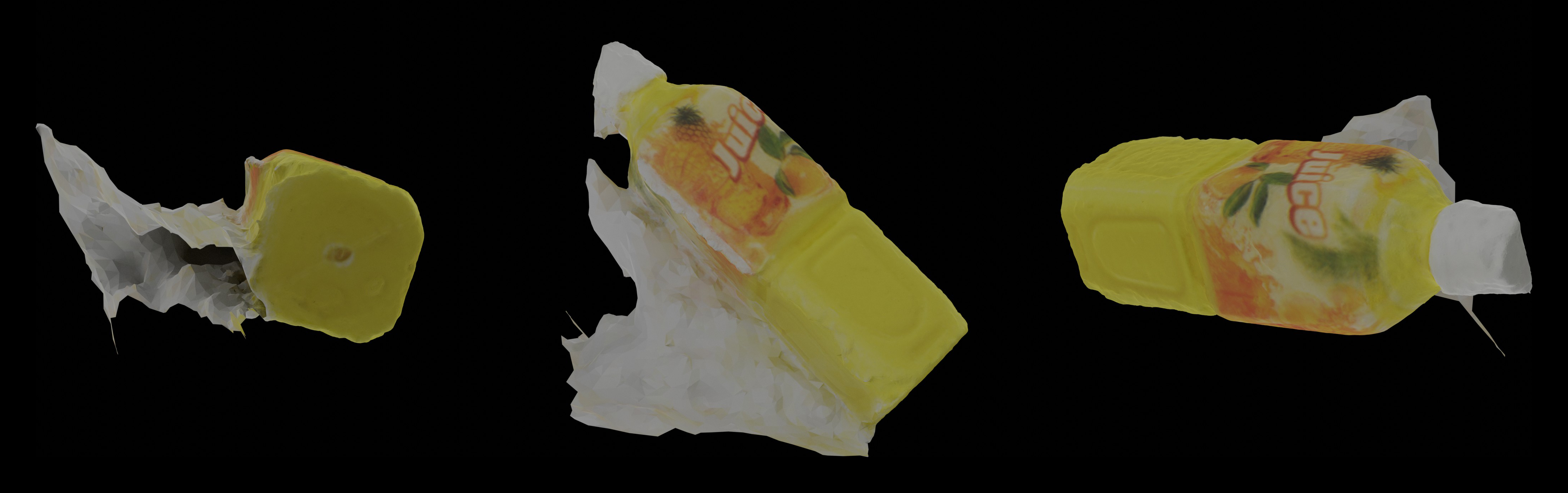}} \\
    \caption{An example of a low-quality reconstruction from three matching perspectives.}\label{fig.global}
    \label{OJ}
\end{figure}

While most of the successfully reconstructed models were of similar quality to the donut and hippo, Figure \ref{OJ} shows an example of a reconstruction of an orange juice bottle which was successful (i.e., returned a model) but is of low quality. This poor performance is likely due to the lack of features on the bottom half of the bottle, which is a solid color, and perhaps relates to the bottle's spacial symmetry, which can cause degeneracies in how feature points are matched. 

As an example of how our pipeline can be used to gain clarity on how certain choices can affect the outcome, we tested different hyperparameters for 8 models which already had base SSIM scores between 0.9 - 0.95. Our base parameters included 100 images for photogrammetric reconstruction, a frame resolution of $2560 \times 1440$ pixels (1440p), and a solid white background.

Our first experiment was to run our pipeline with 70 images and with 130 images, while keeping the resolution, background color, and other hyperparameters the same, to see how it affects the overall SSIM. 

The second experiment was to run our pipeline with a frame resolution of $1920 \times 1080$ pixels (1080p) and with a resolution of $3840 \times 2160$ pixels (4k), while keeping the number of images, background color, and other hyperparameters the same, to see how that affects the SSIM. 

\begin{figure}[htbp]
    \centering
    \includegraphics[width=\linewidth]{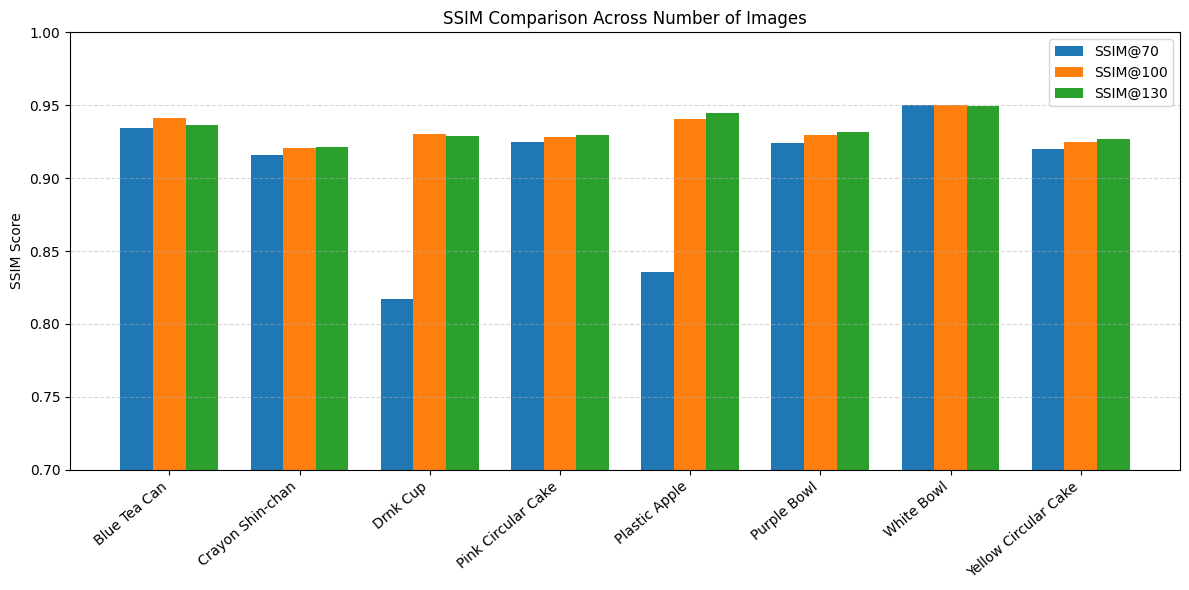}
    \caption{Histogram of results with 70, 100, and 130 frames for reconstruction. The x-axis represents the name of the model, the y-axis represents the SSIM score.}
    \label{fig:numImages}
\end{figure}

\begin{figure}[htbp]
    \centering
    \includegraphics[width=\linewidth]{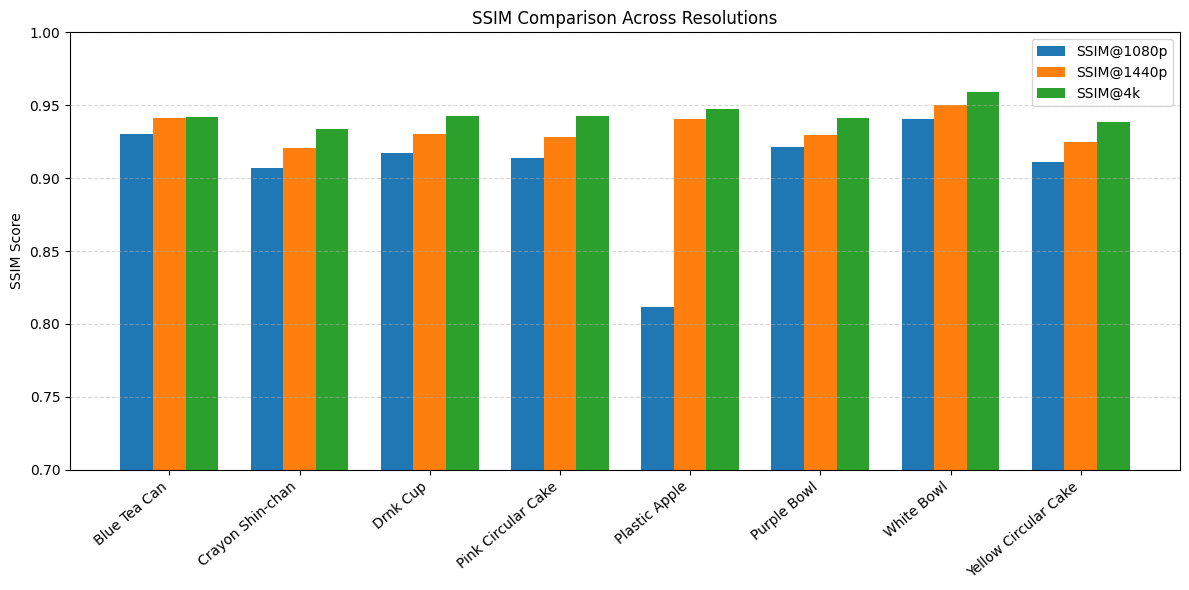}
    \caption{Histogram of results with frames at 1080p, 1440p, and 4k. The x-axis represents the name of the model, the y-axis represents the SSIM score.}
    \label{fig:resolution}
\end{figure}

As figures \ref{fig:numImages} and \ref{fig:resolution} show, increasing the resolution tends to have a larger impact on the SSIM than changing the number of frames, though that can vary based on the size, shape, and surroundings of the object.

\section{Future Work}
\label{sec:futureWork}

The results from our initial run validate the pipeline and open the door for a wide range of systematic, quantitative experiments. Our future work will focus on leveraging this framework to dissect the factors that contribute to digital twin generation and analysis. A major theme will be comprehensive parameter sweeps - i.e., large-scale experiments to isolate the impact of individual parameters on reconstruction quality. These parameters fall under the categories of scene properties, camera configurations, and reconstruction algorithms. Within scene properties, we will systematically vary lighting conditions (e.g., hard vs. soft lighting, number and position of lights) and background properties (e.g., solid colors vs. complex patterns) to quantify their effect on feature matching and texture quality. Camera configuration experiments will entail moving beyond a simple spherical layout to test other camera placement strategies, such as orbital paths and algorithmically-derived ``next best view'' approaches, while varying the number of cameras to map out the cost-benefit curves for different object types (e.g., the orange juice bottle shown in Figure \ref{OJ}). Experiments on reconstruction algorithms will entail comparisons in the performance of different feature detectors (e.g., SIFT, ORB), and evaluating alternative open-source reconstruction back-ends like COLMAP to benchmark their performance on identical input data.

In addition to experiments on parameter choices, the pipeline can be extended to include emerging 3D representations other than meshes, such as Neural Radiance Fields (NeRFs) and 3D Gaussian Splats. This would involve comparing novel view syntheses from the reconstructed models against ground-truth renderings using metrics like PSNR and SSIM.

Finally, the generation of synthetic data from our pipeline, with its attachment to corresponding ground truths, makes it an ideal tool for generating large-scale, highly-annotated datasets for training and validating machine learning models. We plan to create a benchmark dataset with varied lighting, camera poses, and object types to facilitate research in areas like single-shot 3D reconstruction, novel view synthesis, material estimation, and pose estimation.

\section{Conclusion}
\label{sec:conclusion}

In this paper, we presented a novel framework for the quantitative and repeatable analysis of 3D photogrammetry pipelines. By generating synthetic image data from ground-truth 3D models with programmatic control over camera and scene parameters, our methodology enables the direct, empirical measurement of a pipeline's performance. This approach provides a path to move beyond the qualitative and often anecdotal assessments that currently dominate the field.

Our implementation of this pipeline and its application to a large dataset of 3D models successfully demonstrated its viability, highlighting common failure modes which reflect the types of quality degradation that can occur during reconstruction. 

This framework provides the community with a methodology for optimizing complex pipelines, performing direct comparisons between different algorithms, and generating high-quality datasets to train the next generation of 3D vision models. This approach will help establish evidence-based best practices for the creation of high-fidelity digital twins.

{
    \small
    \bibliographystyle{ieeenat_fullname}
    \bibliography{main}
}

\end{document}